\documentclass{article}
\usepackage{PRIMEarxiv}
\usepackage[numbers]{natbib}
\usepackage[utf8]{inputenc} 
\usepackage[T1]{fontenc}    
\usepackage{url}            
\usepackage{booktabs}       
\usepackage{amsfonts}       
\usepackage{nicefrac}       
\usepackage{microtype}      
\usepackage{lipsum}
\usepackage{fancyhdr}       
\usepackage{graphicx}       
\usepackage{multirow}
\usepackage{threeparttable}
\graphicspath{{media/}}  
\usepackage{color,soulutf8}
\sethlcolor{white}
\usepackage{hyperref}
\usepackage{breakurl}
\title{On the benefits of self-taught learning for brain decoding}

\author{
  Elodie Germani \\
  Univ Rennes, Inria, CNRS, Inserm \\
  Rennes, France\\
  \texttt{elodie.germani@irisa.fr} \\
  \href{https://orcid.org/0000-0002-5786-9538}{\includegraphics[scale=0.06]{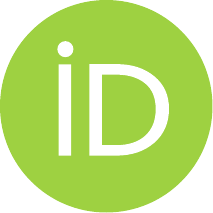}\hspace{1mm} ORCID: 0000-0002-5786-9538} \\
   \And
  Elisa Fromont \footnote[1]{} \\
  Univ Rennes, IUF, Inria, CNRS, IRISA \\
  Rennes, France\\
  \href{https://orcid.org/0000-0003-0133-3491}{\includegraphics[scale=0.06]{orcid.pdf}\hspace{1mm} ORCID: 0000-0003-0133-3491} \\
   \And
  Camille Maumet \thanks{Joint senior authorship}\\
  Univ Rennes, Inria, CNRS, Inserm \\
  Rennes, France\\
  \href{https://orcid.org/0000-0002-6290-553X}{\includegraphics[scale=0.06]{orcid.pdf}\hspace{1mm} ORCID: 0000-0002-6290-553X}
}

\begin{document}
\bibliographystyle{unsrtnat}
\maketitle

\begin{abstract}
\textbf{Context.} We study the benefits of using a large public neuroimaging database composed of fMRI statistic maps, in a self-taught learning framework, for improving brain decoding on new tasks. First, we leverage the NeuroVault database to train, on a selection of relevant statistic maps, a convolutional autoencoder to reconstruct these maps. Then, we use this trained encoder to initialize a supervised convolutional neural network to classify tasks or cognitive processes of unseen statistic maps from large collections of the NeuroVault database. \textbf{Results.} We show that such a self-taught learning process always improves the performance of the classifiers but the magnitude of the benefits strongly depends on the number of samples available both for pre-training and finetuning the models and on the complexity of the targeted downstream task. \textbf{Conclusion.} The pre-trained model improves the classification performance and displays more generalizable features, less sensitive to individual differences.
\end{abstract}


\section{Introduction}
In the past few years, deep learning (DL) approaches have achieved outstanding performance in the field of neuroimaging~\citep{abrol_deep_2021} due to their ability to model complex non-linear relationships in the data. Functional Magnetic Resonance Imaging (fMRI) data, a noninvasive neuroimaging technique in which brain activity is recorded during specific experimental protocols probing different mental processes and giving a big picture on cognition, are often used as input data to these models. 
These can be used for different purpose, such as disease diagnosis \citep{yin_deep_2022} or brain decoding (\textit{i.e.} identifying stimuli and cognitive states from brain activities) \citep{firat_deep_2014}, with a common goal: linking a target with highly variable patterns in the data and ignoring aspects of the data that are unrelated to the learning task. Researchers took advantage of the specific properties of fMRI data to build more and more sophisticated models \citep{vu_3d_2018, hu_multichannel_2019, qureshi_brain_2022, koyamada_construction_2014, wang_decoding_2020, huang_design_2021, vu_fmri_2020, oh_classification_2019}. 

However, training effective DL models using MRI data comes with many challenges \citep{thomas_challenges_2021}. {Brain images differ from natural images that are typically used in the deep learning community on multiple aspects: they contain quantitative information (i.e. statistical values in our context of task-fMRI, that need to be considered differently during preprocessing steps such as standardization or during training with batch normalization), spatial localisation is crucial information (i.e. the same activation in different regions of the brain leads to a completely different interpretation) and the dimensionality of medical images is much larger (i.e. an fMRI statistic map contains tens of thousands of dimensions). This means that a technique used for natural images may fail for medical images} (see \citep{thijs_kooi_deep_2018}).

{While DL models have helped resolve important problems in sub-fields of brain imaging (such as the segmentation of anatomical datasets) their use in task-fMRI is still limited} \citep{sun_intelligence_2022}. {Performance of DL models in task-fMRI has been limited by the high dimensionality, lack of diversity and low sample size of conventional datasets} \citep{poldrack_scanning_2017, button_power_2013}.  {In relation with the large number of trainable parameters in DL models}~\citep{cho_how_2016}{, this makes it particularly difficult to build fair and generalizable DL models for fMRI. Indeed, fMRI datasets are typically composed of 3D volumes with hundred thousand dimensions (or voxels) for a rather small number of subjects (typically 10-100). The field also suffers from a large number of sources of variability in the data at the subject level (brain activity patterns differ across subjects), the acquisition level (fMRI scanners and protocols often vary between centers and studies) and the analysis level (different analysis pipelines lead to different brain patterns). In our case, brain decoding models should be robust to all these sources of variability} \citep{ricci_lara_addressing_2022}.

This problem of small and uniform training sets is not limited to neuroimaging and is well known in the field of machine learning, where researchers extensively use deep transfer learning to improve classification and generalization performance of their models (see for example \citep{pan_survey_2010}). This method consists in using the knowledge obtained from a model trained for a source task on a source dataset and applying it to a target task on a target dataset. Transfer learning proved its worth on natural images by using large, publicly available datasets \citep{deng_imagenet_2009} to pre-train DL models before fine-tuning them on smaller datasets of a related domain. 

{While the availability of large datasets has enabled solving difficult machine learning problems (such as classification), collecting similarly large datasets in brain imaging (including a diversity of participants and modalities) is especially challenging. In fact, in fMRI studies, the median sample size was still limited to N=30 participants in 2015} \citep{poldrack_scanning_2017}. {Efforts for collecting large scale datasets have arisen in the field in the past 10 years with for instance the Human Connectome Project (HCP)} \citep{glasser_human_2016} {or the UK Biobank} \citep{sudlow_uk_2015} { and give hope for an improvement of model performance. But these datasets are still much smaller than those that brought breakthroughs in computer vision and are often much less diverse}~\citep{ge_increasing_2023}.

To prevent overfitting and allow for generalizable statistical inference, neuroimaging researchers proposed methods to tackle this lack of training data \citep{bontonou_few-shot_2021, yotsutsuji_evaluation_2021, zhuang_fmri_2019}. For instance, \cite{mensch_extracting_2021} built a decoding model using data gathered from 35 studies and thousands of individuals that cover various cognitive domains. Despite the good performance of the models, these can only be applied on restricted sets of studies, discriminating between few cognitive concepts. More annotated training data (\textit{e.g.} using large public databases) would be required to map a wider set of cognitive processes. 

Lots of studies were also made on inductive transfer learning with labeled source data as defined in \cite{pan_survey_2010} (e.g. source task and target task are different, as well as source domain and target domain) \citep{thomas_evaluating_2021, y_gao_decoding_2019, svanera_transfer_2019}. For instance, \cite{thomas_evaluating_2021} pre-trained two DL classifiers on a large, public whole-brain fMRI dataset of the HCP, fine-tuned them and evaluated their performance on another task on the same dataset and on a fully independent dataset. In another study, \cite{y_gao_decoding_2019} used the ImageNet database \citep{deng_imagenet_2009}, a large, public dataset containing naturalistic images from more than 1000 classes, to pre-train a model and adapt it to classify tasks from 2D fMRI data. This database was also used in \cite{MALIK2022325} for pre-training a 2D structural MRI classifier. In the same paper, the Kinetics dataset~\citep{kay2017kinetics} was also used to evaluate the transfer learning process with 3D images. In a recent work, \cite{thomas2022self} used self-supervised learning frameworks to pre-train DL decoding models across a broad fMRI dataset, comprising many individuals,
experimental domains, and acquisition sites. These studies show improved classification accuracies as well as quicker learning and less training data required. 

However, labeled databases are not always available in neuroimaging, despite the growing effort in data sharing to build public databases \citep{poldrack_making_2014}, such as OpenNeuro for raw data \citep{markiewicz_openneuro_2021} and NeuroVault for fMRI statistic maps \citep{gorgolewski_neurovaultorg_2015}. The unconstrained annotations and the heterogeneity of tasks and studies make them difficult to use to pre-train a supervised deep learning model. To compensate this, weakly supervised learning techniques such as automatic labelling of data has proven its worth. For instance, \cite{menuet_comprehensive_2022} enriched NeuroVault annotations using the Cognitive Atlas ontology \citep{poldrack_cognitive_2011} and used these labeled data to train a multi-task decoding model that successfully decoded more than 50 classes of mental processes on a large test set. 

A specific type of inductive transfer learning named {\it self-taught learning}
\citep{raina_self-taught_2007, wang_robust_2013} showed strong empirical success in the field of machine learning. It does not require any labels as it consists in training models to autonomously learn latent representations of the data and using these to improve learning in a supervised setting. This approach is motivated by the observation that data from similar domains contain patterns that are similar to those of the target domain. By initializing the weights of a supervised classifier with the pre-trained weights of an unsupervised model trained on many images. The aim is to improve the model performance by placing the parameters close to a local minimum of the loss function and by acting as a regularizer \citep{erhan_why_nodate}. 

In the field of neuroimaging, latent representations have recently been used in a task-relevant autoencoding framework. \cite{orouji2022task} used an autoencoder with a classifier attached to the bottleneck layer on a small fMRI dataset. This model outperformed the classifier trained on raw input data by focusing on cleaner, task-relevant representations. This suggests that a low-level representation of fMRI data, learned for a reconstruction task, can be helpful in a classification task, as in a self-taught learning framework.

In this work, we propose to take advantage of NeuroVault -- a large public neuroimaging database {that was built collaboratively and therefore displays a good level of variability in terms of fMRI acquisition protocols, machines, sites and analysis pipelines} -- in a self-taught learning framework. We pre-trained an unsupervised deep learning model to learn a latent representation of fMRI statistic maps and we fine-tune this model to decode tasks or mental processes involved in several studies. In a first part, we leveraged the NeuroVault database to select the most relevant statistic maps and train a convolutional autoencoder (CAE) to reconstruct these maps. In a second part, we used the final weights of the encoder to initialize a supervised convolutional neural network (CNN) to classify the cognitive processes, tasks or constrasts of unseen statistic maps from large collections of the NeuroVault database (an homogeneous collection of more than 18000 statistic maps and an heterogeneous one with 6500 maps). Our goal was to investigate how {the use of a large and diverse database in a} self-taught learning {framework} can be beneficial in the field of brain imaging for deep learning models. 

\section{Material \& Methods}
The code produced to run the experiments and to create the figures and tables of this paper is available in the Software Heritage public archive \cite{code_SH}. Derived data used by these notebooks are stored in Zenodo \citep{derived_data}.

\begin{figure*}
	\centering
		\includegraphics[width=0.99\textwidth]{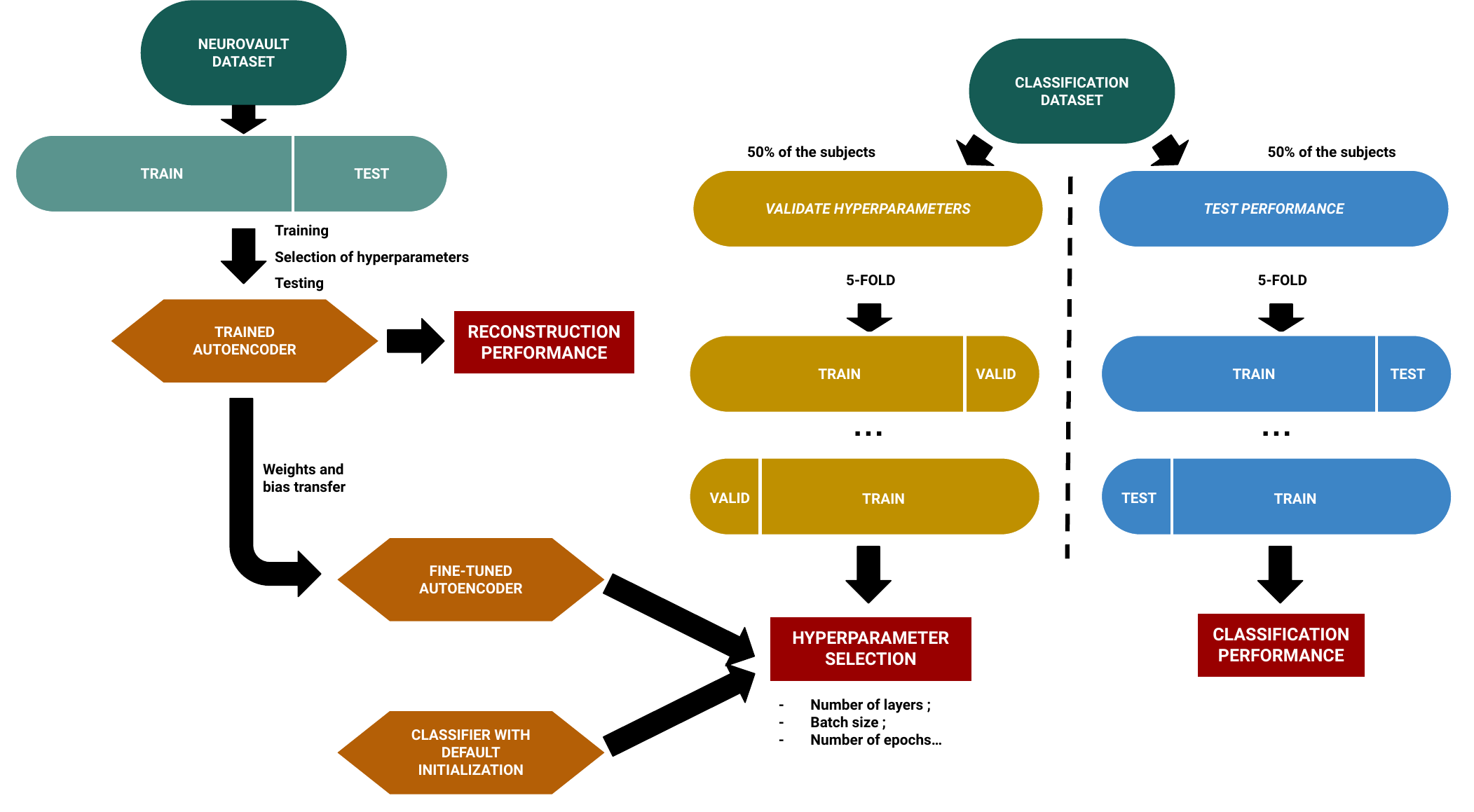}
	\caption{\hl{Flow diagram of the self-taught learning methodology. NeuroVault dataset is used to train a Convolutional AutoEncoder (CAE). The encoder of this CAE is used to initialize a Convolutional Neural Network (CNN) and to train it to classify other datasets. These classification datasets are split in two disjoints datasets: a `validation' one used to optimize hyperparameters and a `test' one to evaluate performance. In each one, a 5-fold cross-validation is performed.}}
	\label{fig:fig1}
\end{figure*}

Figure \ref{fig:fig1} illustrates the overall process used to implement our self-taught learning framework: a CAE was first trained to reconstruct the maps of a large dataset extracted from NeuroVault. Then, the encoder part of the CAE was fine-tuned to answer a classification problem on another dataset (with labels). After hyperparameters optimisation, performance of the pre-trained classifier was compared to those of a classifier initialized with a default algorithm. Details regarding the datasets (NeuroVault dataset and classification datasets) can be found in the next subsection. The models of the CAEs and the CNNs are presented in Section \ref{model_architecture}. Further explanations on the workflow used to train the CAE and the CNN and to evaluate their performance are available in Sections \ref{cae_training} and \ref{cnn_training} respectively. 

\subsection{Overview of the datasets}

A summary of the different datasets can be found in Table \ref{tab:tab1}. Details are given below. 

\begin{table}[h]
	\centering
	\caption{Overview of the datasets. For each dataset, number of statistic maps are presented, as well as the number of subjects, number of studies and the type of labels (if available).}\label{tab:tab1}
	\footnotesize
		\begin{tabular}{ccccc}
		\hline
		\centering \textbf{Dataset} & \centering \textbf{Maps} & \centering \textbf{Subjects} & \centering \textbf{Studies} & \centering \textbf{Labels}
		\tabularnewline 
		\hline
		NeuroVault & 28,532 & - & - & -
		\tabularnewline 
		& & & &
		\tabularnewline 
            \hline
		HCP & 18,070 & 787 & 1 & Tasks (\textit{7})
		\tabularnewline 
		& & & & Contrasts (\textit{23})
		\tabularnewline
		& & & &
		\tabularnewline 
            \hline
		BrainPedia & 6,448 & 826 & 29 & Cognitive
		\tabularnewline
		& & & & processes (\textit{36})
		\tabularnewline
		\end{tabular}
\end{table}

\subsubsection{NeuroVault dataset} \label{neurovault_dataset}
NeuroVault \citep{gorgolewski_neurovaultorg_2015} (RRID:SCR$\_$003806) is a web-based repository for statistic maps, parcellations and atlases produced by MRI and PET studies. This is currently the largest public database of fMRI statistic maps. 
NeuroVault has its own public Application Programming Interface (API) that provides a full access to all images (grouped by collections) and enables filtering of images or collections with associated metadata. At the time of experiment (19/01/2022), a total of 461,461 images in 6,782 collections were available. 
Among the available metadata, some are mandatory and specified for all maps such as the modality (e.g. "fMRI-BOLD" for Blood-Oxygen Level Dependent Functional MRI; dMRI for Diffusion MRI, etc.), the type of statistic  (e.g. "T map" or "Z map") or the cognitive paradigm (e.g. "Working memory" or "Motor fMRI task paradigm"), and others are optional and only available if additionally entered at the time of the upload. 

From this large database, relevant maps were selected based on multiple criteria. First, we chose maps for which the modality was `fMRI-BOLD' to exclude other modalities such as structural or diffusion MRI. To get comparable maps, we set three additional inclusion criteria and selected maps: 1/ for which all required metadata were provided (`is$\_$valid' to True) 2/ that were registered in MNI space (`not$\_$mni' to False) -- to ensure that anatomical structures were located at the same coordinates in each map -- and 3/ referenced as `T map' or `Z map' -- to exclude maps in which voxel values did not have the same meaning (e.g. P value maps, Chi-squared maps, etc.) --. Among these, thresholded statistic maps were excluded. 

We found that some maps in our initial dataset, were wrongly referenced as T map or Z map. These misclassified maps were removed by filtering the `filename' column of the dataframe to exclude \textit{SetA$\_$mean} \textit{SetB$\_$mean} (AFNI contrast maps), \textit{con} (SPM contrast maps), \textit{cope} (FSL contrast maps).

Using these criteria, a total of 28,532 statistic maps were selected from the NeuroVault database and constituted our `NeuroVault dataset'. Most of these maps were unlabeled (\textit{i.e.} cognitive processes or tasks performed described as `None / Other') or not labeled in a standardized way (\textit{i.e.} use of terms that are specific for a study instead of generic terms, such as those defined in ~\cite{poldrack_cognitive_2011} \textit{e.g.} some maps were labeled as `word-picture matching task' for the cognitive paradigm whereas others in which a similar task was performed were referenced as `working memory fMRI task paradigm' which is a label that includes other specific tasks).

\subsubsection{HCP dataset (NeuroVault Collection 4337)} \label{hcp_dataset}
NeuroVault collection 4337~\cite{neurovault_collection_4337} includes 18,070 z-statistic maps, for base contrasts (task vs baseline), corresponding to 787 subjects of the HCP~900 release \citep{van_essen_wu-minn_2013}. This collection was excluded from our pre-training dataset (see section \ref{neurovault_dataset}) due to missing metadata (i.e. `is$\_$valid' is False). 

All maps in this collection were grouped together and referred to as the `HCP dataset' in the following. Multiple labels were entered for each map including:  mental concepts (`cognitive\_paradigm\_cogatlas'), tasks (`task') and contrasts (`contrast\_definition') (as defined in ~\cite{poldrack_cognitive_2011}). For each subject, 23 contrasts distributed in 7 tasks were available: 
\begin{itemize}
    \item Working memory: `0-back body', `0-back face', `0-back places', `0-back tools', `2-back body', `2-back face', `2-back places', `2-back tools'
    \item Motors: `cue', `left foot', `left hand', `right foot', `right hand'
    \item Relational: `relational', `match'
    \item Gambling: `punish', `reward'
    \item Emotion: `faces', `shapes'
    \item Language: `math', `story'
    \item Social: `tom'
\end{itemize} 
For more details on contrasts, tasks and mental concepts of this study, see \cite{van_essen_wu-minn_2013}. 

\subsubsection{BrainPedia dataset (NeuroVault collection 1952)}
NeuroVault collection 1952~\cite{neurovault_collection_1952}, known as BrainPedia~\citep{varoquaux_atlases_2018}, contains fMRI statistic maps of about 30 fMRI studies from OpenNeuro \citep{markiewicz_openneuro_2021}, the Human Connectome Project \citep{van_essen_wu-minn_2013} and from data acquired at Neurospin research center, together they were chosen to map a wide set of cognitive functions. 

This collection contains 6,573 statistic maps corresponding to 45 unique mental concepts derived from 19 sub-terms (\textit{e.g.} `visual, right hand, faces' for maps associated with the task of watching an image of a face and responding to a working memory task). These images were previously used to build a multi-class decoding model \citep{varoquaux_atlases_2018} and labels corresponded to the mental concepts associated with the statistic map, e.g., `visual', `language' or `objects'. Here we excluded the nine classes that had less than 30 samples each, leaving 6,448 images corresponding to 36 classes. These 6,448 images were grouped together and referred to as the `BrainPedia' dataset in the following.

\subsection{Preprocessing}
All statistic maps included in this study were downloaded from different collections of NeuroVault and therefore were processed using different pipelines (see the original studies for more details \cite{varoquaux_atlases_2018}, \cite{van_essen_wu-minn_2013}).
We resampled all maps to dimensions (48, 56, 48) using the MNI152 template available in Nilearn \citep{abraham_machine_2014} (RRID: SCR\_001362) as target image. A min-max normalization was also performed on all resampled maps to get statistical values between -1 and 1. Finally, the brain mask of the MNI152 template in Nilearn was used to exclude statistical values outside the brain in all statistic maps. 

\subsection{Model architectures} \label{model_architecture}

\begin{figure*}
	\centering
		\includegraphics[width=0.99\textwidth]{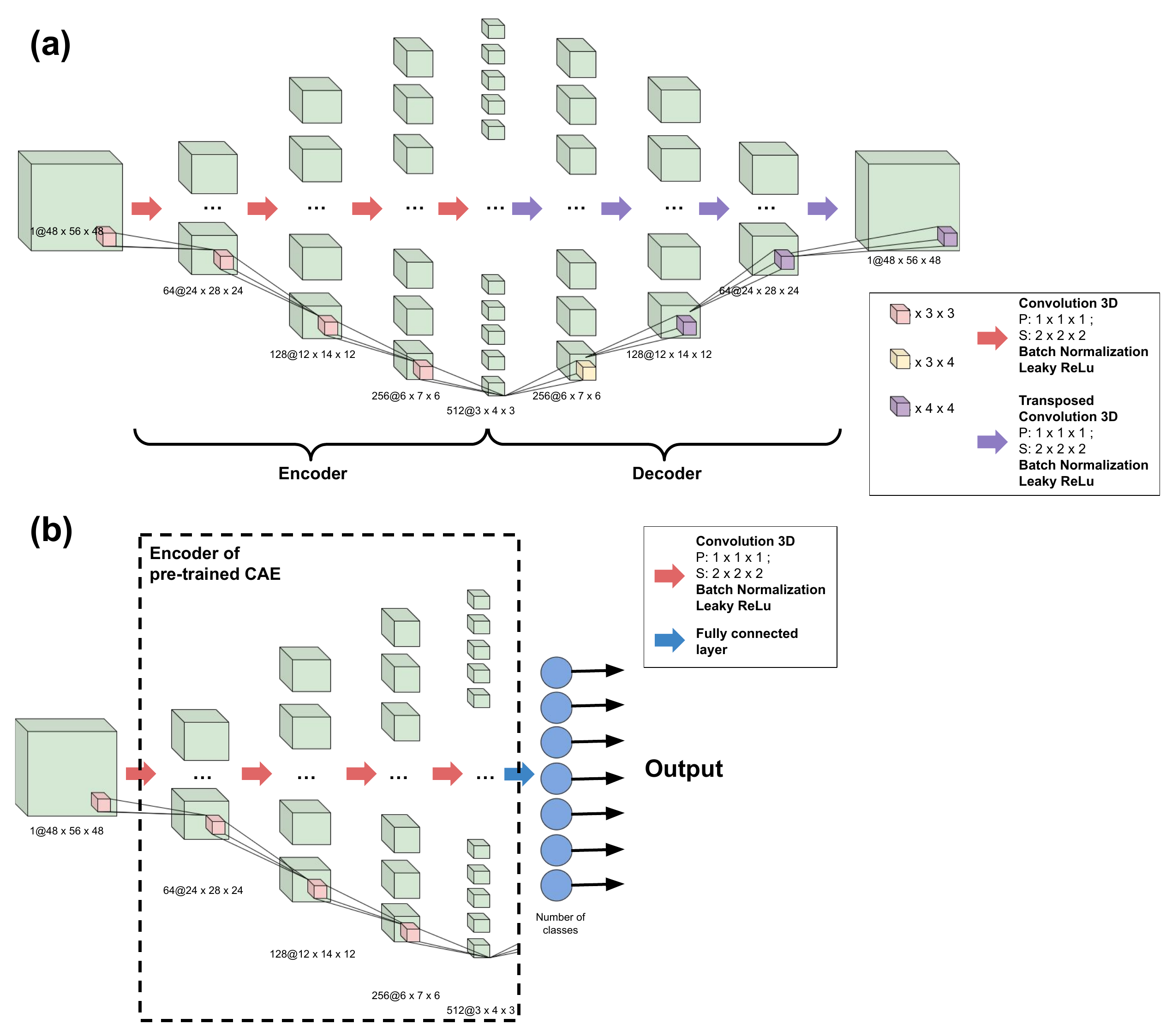}
	\caption{\hl{Schematic visualisation of the architectures of the Convolutional AutoEncoder (a) and Convolutional Neural Network (b) with 4 layers. The CAE is composed of an encoder and a decoder with respectively 4 convolutional and transposed convolutional layers. The size of the latent space is 512 * 3 * 4 * 3. The CNN has the same architecture as the encoder of the CAE with a fully-connected layer added at the end of the network with different numbers of output node depending on the dataset and the classification performed.}}
	\label{fig:fig2}
\end{figure*}

All models were implemented using PyTorch~\citep{pytorch_2019} v1.12.0 (RRID:SCR\_018536) with CUDA~\citep{cuda_2012} v10.2. For our model architectures, we chose to use 3D-convolutional feature extractors that take into account the three spatial dimensions of fMRI statistic maps. Schematic representations of the architectures are available in Figure \ref{fig:fig2} and Supplementary Figure~S1, available at \citep{supporting_data}. 

\subsubsection{Convolutional AutoEncoder (CAE)}
The base architecture of our CAE was inspired from \cite{zhuang_fmri_2019}. Two architectures were derived from this base: a 4 layers and a 5 layers architecture, respectively corresponding to the number of convolutional layers in each part of the CAE (encoder and decoder). In the 4-layer model, the encoder part consisted in four 3D convolutional layers with respectively 64, 128, 256 and 512 channels. Each layer had a kernel size of 3~x~3~x~3, a stride of 2~x~2~x~2 and a padding of 1~x~1~x~1. 3D batch normalization layers \citep{batch_norm} followed each convolutional layers with respectively 64, 128, 256 and 512 channels and a leaky rectified linear unit (ReLU) activation function was used for all layers. The decoding part of the CAE was symmetric to the encoder, except that 3D transposed convolutional layers were used instead of classic convolutional layers. Transposing convolutions is a method to upsample an output using learnable parameters. It can be seen as an opposite process to classical convolutions. To keep the number of features symmetric at each layers output, the kernel size of the first layer was set to 4~x~3~x~4 and to 4~x~4~x~4 for all other transposed convolutional layers. Leaky ReLU activation function was also used for all layers except for the last one, \textit{i.e.} the output one, for which a sigmoid function was used in order to obtain output values between -1 and 1. The latent space for this model was of size 512 x 3 x 4 x 3. A schematic representation of this architecture can be found in Figure \ref{fig:fig2}(a).

In the 5-layer model, one convolutional layer was added at the beginning of the encoder with 32 channels and similar parameters as the other layers of the encoder. A transposed convolutional layer was also added at the end of the decoder with 32 channels. The kernel sizes in the decoder were also modified to maintain the feature map sizes: the first and second layers of the decoder had kernel sizes of 3 x 4 x 3 and 4 x 3 x 4 respectively. All other parameters, batch normalization layers and activation functions were the same. The latent space for this model was of size 512 x 2 x 2 x 2. A schematic representation of this architecture can be found in the Supplementary Figure S1 (a), available at \citep{supporting_data}.

\subsubsection{Convolutional Neural Network (CNN)}
The 3D CNNs used for classification followed the architecture of the encoder part of the CAEs. In the same way as for the CAEs, two CNN architectures were derived. For each one, we took the corresponding architecture of the encoder (4 or 5 layers) and added a fully connected layer at the end. The number of nodes in this layer varied depending on the number of classes. A softmax activation function was used for this output layer. Visual representation of the CNNs are available in Figure \ref{fig:fig2}(b) and Supplementary Figures S1 (b), available at \citep{supporting_data}.

\subsection{CAE training} \label{cae_training}

To train our CAEs to reconstruct the statistic maps of the NeuroVault dataset, we used an Adam optimizer \citep{kingma2014adam} with a learning rate of 1e-04 and all other parameters with default values. The loss function was the Mean Squared Error (MSE: the squared L2 norm) which is the standard reconstruction loss.

\subsubsection{Dataset split}
NeuroVault dataset was randomly split in two subsets: training and test with respectively 80\% and 20\% of the maps. The training set (N=22,772 maps) was used to train the CAE with the different architectures and the test set (N=5,760 maps) to assess the performance of the different models (with different hyperparameters).

\subsubsection{Architecture comparison}
To limit the computational cost of our experiments, we fixed some of the hyperparameters of the CAE and only compared those who were of interest for the later experiments. Here, we use the term model “hyperparameters”, to distinguish with model “parameters”, to represent the values that cannot be learnt during training, but are set beforehand e.g., the batch size or the number of hidden layers. Thus, a batch size of 32 and a learning rate of 1e-04 were chosen to train the CAE for a number of 200 epochs (\textit{i.e.} values that are often used in experiments). The only hyperparameter for which different values were compared were the number of hidden layers of the model: 4 layers vs 5 layers for each part (encoder/decoder) of the model.

\subsubsection{Performance evaluation}
To assess the performance of the CAEs, we estimated Pearson's correlation coefficient between the reconstructed statistic map and the original statistic map. The correlation coefficient was computed using numpy version 1.21.2 (RRID: SCR\_008633)\citep{harris_array_2020}. The closer to 1 the correlation coefficient was, the stronger the relationship between the maps and the more accurate the reconstruction. Note that we did not use MSE in this context as its individual values (for each data point) were not easily interpreted. 

\subsection{Classifier training} \label{cnn_training}

We trained two types of classifiers for all the experiments: 
\begin{itemize}
    \item the \textit{classifier with default algorithm} initialized with the original algorithm from~\citep{he2015delving} (\textit{i.e.} Kaiming Uniform algorithm for convolutional and fully-connected layers with a parameter of {$\sqrt{5}$}) and \item the \textit{classifier with pre-trained CAE }  initialized using the weights and bias of the convolutional layers of the CAE pre-trained on NeuroVault dataset.
\end{itemize}

 The CNNs were trained using the Adam optimizer with a learning rate of 1e-04. We used the cross-entropy loss function for training the classifier. Both were implemented in PyTorch.

\subsubsection{Dataset split}

{As described in Fig.~{\ref{fig:fig1}} (on the right), the classification datasets were split in two disjoint subsets: the \textit{`validation dataset'} used to optimize the hyperparameters, and the \textit{`test dataset'} used to test the performance. Each subset contained 50\% of the subjects of the overall dataset with no overlap to avoid any data leakage (see \mbox{\citep{varoquaux_machine_2022, data_leakage_ML}}).}

For each experiment, the validation and test datasets were then split into 5 folds for cross-validation. Subjects were randomly sampled in each fold in order to ensure that there was no overlap of subjects across folds. The identifiers of the subjects included in the different folds were saved for reproducibility. More details on the methods used to perform the 5-folds split for each dataset are specified in subsection \ref{cv-split}.  

\subsubsection{Evaluation of performance}
The performance of each model was measured using several metrics: accuracy (Acc), precision (P), recall (R) and F1-macro score (F1). 
All metrics were implemented using scikit-learn \citep{abraham_machine_2014} with default parameters, except for F1-score for which the "average" parameter was specified with "macro" to deal with multi-class classification. 

To evaluate the performance of a model, all metrics were averaged among the 5 folds of cross-validation and standard error of the mean was computed. 

To compare the final performance of models with default initialization versus fine-tuned weights, we used paired {one-tailed} two-sample t-tests between the performance values (accuracy or F1-score) of the 5 models trained during cross-validation. T-statistic and p-value were provided and value of 0.05 was used for the p-value significance threshold. 

\subsubsection{Hyperparameters optimisation}
{To select the best hyperparameters for each dataset and each type of initialization, we evaluated the performance of each model by performing a 5-fold cross-validation on the validation dataset.}

For each type of classifier (i.e. initialized with default algorithm versus pre-trained), we refined and optimised the hyperparameters using the largest datasets (Large BrainPedia and HCP). However, the large amount of training data made it computationally extremely costly to perform a full grid-search. We therefore limited our research to predefined values of batch sizes (32 or 64), number of epochs (200 or 500) and model architectures (4 layers or 5 layers). {All batch sizes, number of epochs and architectures were tested for each type of classifier and each dataset. We didn't perform any optimization on the learning rate to limit the computational cost of our experiments. Every model was trained using a learning rate of 1e-04.}

{We selected the best set of hyperparameters based on the performance of the corresponding model in terms of accuracy and F1-score, averaged across folds.} 

\subsection{Benefits of self-taught learning and impact of different factors} \label{cv-split}
To investigate the benefits of self-taught learning for neuroimaging data, different brain decoding experiments were studied. {For all, after optimizing the hyperparameters of the two models (i.e. the model with default initialization -or- with pre-trained CAE and fine-tuned weights) we assessed the performance of these optimized models on the test dataset using a 5-fold cross-validation.}

\begin{figure*}
	\centering
		\includegraphics[width=0.99\textwidth]{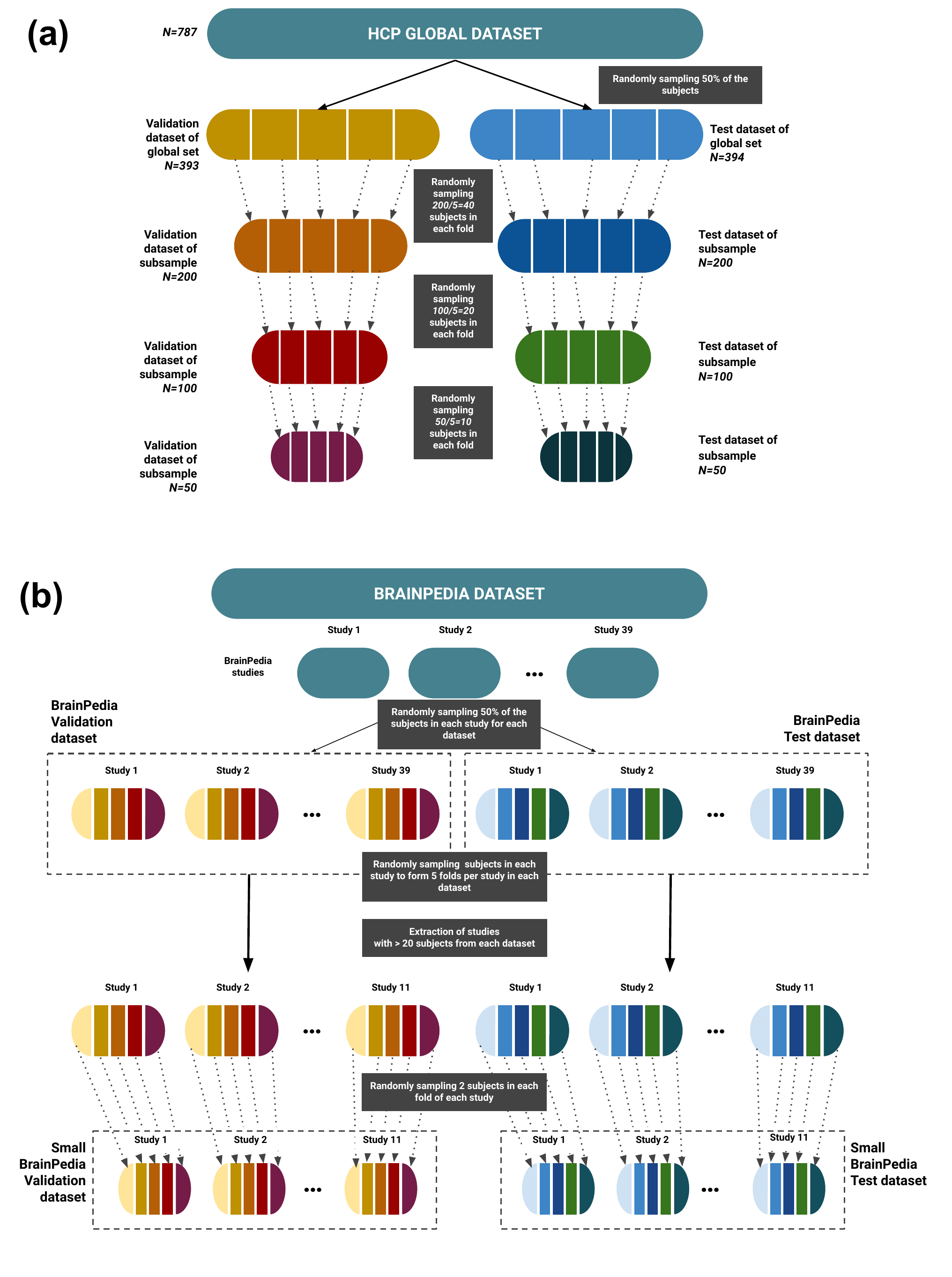}
	\caption{Overview of the process used to split the datasets for cross-validation. (a) shows the method performed for HCP dataset and its subsamples and the one used for BrainPedia and Small BrainPedia datasets is presented in part (b). In both cases, the global dataset is first split into two subdatasets `validation' and `test' with respectively 50\% of the subjects and then each subdataset is divided into 5 folds for cross-validation.}
	\label{fig:fig3}
\end{figure*}

\subsubsection{Homogeneous dataset (single study)} \label{hcp_classif}
The HCP dataset was used to compare the performance of the models for the task of decoding on a homogeneous dataset (\textit{i.e.} from a single study). We studied the impact of two factors on the classification: sample size and number of target classes. For sample size, subsets of the global HCP dataset were created with different number of subjects: N=50, 100 and 200. Each smaller subset being a subset of the immediately larger one. {To create these subsets, we first split the global HCP test dataset into 5 folds, with different subjects in each fold. In each of these 5 folds, we randomly sampled $200/5 = 40$ subjects and obtained 5 sub-folds that together composed the smaller subset of 200 subjects. This process was repeated for subsamples N=100 and 50 by sampling from their superset. This insured that the 5 models trained on different combinations of the 4 folds of a smaller subset could be tested on the remaining fold of the global test dataset with no overlap between the training and test data. The process is illustrated in Fig.~\mbox{\ref{fig:fig3}(a)}.}

{In the end, we obtained 4 datasets with respectively N=50, 100 and 200 subjects in addition to the global dataset with all subjects (N=393). These datasets respectively contained 1150, 2300, 4590 and 9017 statistic maps in the test subset and 1150, 2300, 4591 and 9053 in the validation subset (note: some contrasts were missing for part of subjects). Since we use a 5-fold validation scheme, the models were trained on approximately $80\%$ of the statistic maps in the corresponding subset (\textit{i.e.} validation for hyperparameter optimization and test for performance evaluation).}

Three types of classification were investigated. First, the `contrast classification' which consisted in identifying the contrast associated with a statistic map (23 different contrasts). Second, the `task classification' which consisted in identifying the task associated with a statistic map (7 different tasks, with multiple contrasts per task). Third, the `one contrast task classification'. This time, we selected a single contrast per task and classified the tasks (7 different tasks, with one contrast per task). The selected contrasts were `2-back places', `faces', `punish', `relational', `right hand', `story' and `tom' respectively for the tasks `Working Memory', `Emotion', `Gambling', `Relational', `Motor', `Language', 'Social'. {We selected these contrasts similarly to what was done in} \cite{wang_decoding_2020} {in which the HCP dataset was used in a decoding model. For each task, the contrast that showed a greater association with the task had priority over the other (for instance, `punish’ for the `Gambling’ task). For `Working Memory’ and `Motor’ tasks, which contained more than one task condition, they randomly chose one (`2-back body' for Working Memory and `right hand' for Motor).} {The dataset used for this third type of classification was thus smaller than the others (only one map per task per subject)}. {For this classification task, the number of statistic maps was respectively 300, 598, 1198 and 2355 for N=50, 100, 200 and for the global dataset.}

\subsubsection{Heterogeneous dataset (multiple studies)}
To study the benefits of self-taught learning on a heterogeneous dataset (i.e.\ from multiple studies), we used BrainPedia. For these experiments, we focused on the classification of mental concepts (as available in NeuroVault metadata). Fig.~\ref{fig:fig3}(b) illustrates the process used to split this dataset. To perform the split while maintaining the heterogeneity in each fold, we randomly sampled 50\% of the subjects of each study to form the `validation' and `test' datasets of BrainPedia (see Fig~\ref{fig:fig3} b.). Then, each dataset, each study was split into 5-folds and the n-th folds of the different studies were combined to form the n-th fold of the dataset. {Validation and test datasets included N=428 subjects and were respectively composed of 3179 and 3269 statistic maps.}

We also studied the impact of sample size in the presence of heterogeneity by extracting smaller datasets. Among the 29 studies of the BrainPedia dataset, we only kept those which were composed of more than 20 subjects. In these remaining studies, already split into 5 folds in BrainPedia validation and test subdatasets, 2 subjects were randomly drawn per fold per study per subdataset to obtain 10 subjects per study per subdataset. Like above, the n-th folds of the different studies were combined to form the n-th fold of each subdataset of the `Small BrainPedia' dataset. In the end, this smaller dataset was composed of 1,844 maps, divided in 30 classes, from 11 studies and 220 subjects. {This dataset was also split into test and validation subsets with 50\% of the subjects in each (N=110). The test and validation subsets were thus composed respectively of 917 and 927 maps.}

\subsection{Explainability}

\subsubsection{Exploring feature maps to understand the generalizability across subjects}
{To investigate the reasons for the difference in performance between the pre-trained and default models, we visualized and analyzed the feature maps of the different convolutional layers of the model. Visualizing these features was useful to better understand how each model made its predictions. Differences in the observed features can help understand differences in terms of performance.}

{With a generalizable classifier, we hypothesized that features of different subjects from the same class should be similar (and therefore not be impacted by individual differences). To study this, we computed for each classifier, each layer and each class, the correlations between the feature maps for all pairs of subjects. A high mean correlation highlighted a higher similarity between the feature maps extracted by this layer for a classifier and thus a higher generalizability.}

\subsubsection{Investigating the contribution of each layer to the overall performance}
{We explored which pre-trained layer had the strongest impact on the classification performance. This could be made at two stages: before and during training. 

Before training, we only \hl{transferred} the weights of some parts of the CAE. In particular, we kept the weights of the last convolutional layers with a default initialization and initialized the first layers with the weights of the pre-trained CAE. Multiple configurations were explored: \hl{transferring only the weights of the first one up to the first four convolutional layers}.} 

{During training, we froze some layers of the model initialized with the weights of the pre-trained CAE, i.e. some layers (the first ones) were not fine-tuned. Multiple types of freezing were tested: \hl{freezing of the first two to the first five convolutional layers}.}

\section{Results}
\subsection{AutoEncoder performance}
\begin{figure*}
	\centering
		\includegraphics[width=0.9\textwidth]{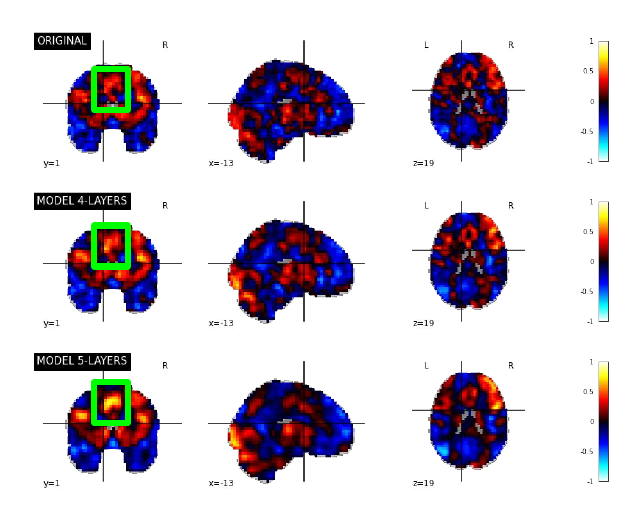}
	\caption{Original version and reconstruction of a randomly drawn statistic map of NeuroVault test dataset (image ID: 109) with the two CAEs (4-layers and 5-layers). {The green square corresponds to a highlighted part of the map for which reconstruction performance are better using the 5-layers architecture.}}
	\label{fig:fig4}
\end{figure*}

Reconstruction performance of the CAEs is presented in Table \ref{tab:tab2}. When comparing the two CAE architectures (4-layers vs 5-layers) trained on NeuroVault dataset, the mean correlations between original and reconstructed maps were better for the 4-layers architecture (86.9\% vs 77.8\%). These results suggest that the reconstruction capabilities of the CAEs are dependant on the model architecture and the size of the latent space. Figure \ref{fig:fig4} shows the reconstruction of a statistic map randomly drawn from the NeuroVault test dataset with the two CAE architectures. With the 4-layers architecture, details of the map were better reconstructed than with the 5-layers architecture (see the green square on the map). This was due to the level of compression of the data that was higher in the 5-layers CAE and that learned only the most useful features with less emphasis in learning specific details. Both models were used as pre-trained model for classification to see if the benefits of the CAEs were related to their reconstruction performance.

\begin{table}[h]
	\centering
	\caption{Reconstruction performance of the CAE depending on model architecture and training set. Values are the mean Pearson's correlation coefficients (standard error of the mean).}\label{tab:tab2}
	\footnotesize
		\begin{tabular}{c|c|c}
		\centering \textbf{Model} & \textbf{4-layers} & \textbf{5-layers}
		\tabularnewline 
		& \textit{Latent space 18,432} & \textit{Latent space 4,096}
		\tabularnewline
		\hline
		\centering Correlation & 86.9 & 77.8 
		\tabularnewline
		(\textit{std error}) & (\textit{0.18}) & (\textit{0.23})
		\end{tabular}
\end{table}

\subsection{Hyperparameters optimisation for classifiers}
The best hyperparameters and corresponding performance can be found on Table \ref{tab:tab3}.

{\renewcommand{\arraystretch}{2}
\begin{table*}[h]
	\centering
	\caption{{Hyperparameters chosen for each dataset and corresponding performance of the classifier on the validation set of the dataset.}}\label{tab:tab3}
	\footnotesize
		\begin{tabular}{c|c|c|c|c|c|c}
		\centering \textbf{Dataset} & \textbf{Initialization} & \textbf{Model} & \textbf{Epochs} & \textbf{Batch} & \textbf{Average Accuracy (\%)} & \textbf{Average F1 (\%)}
		\tabularnewline 
            \centering & & & & & (\textit{std. err.}) & (\textit{std. err.})
            \tabularnewline
		\hline 
		\centering HCP & Default algorithm & 4-layers & 500 & 32 & 90.8 (\textit{1.5}) & 90.8 (\textit{1.6})
            \tabularnewline
		\centering & Pre-trained CAE & 5-layers & 200 & 64 & 91.8 (\textit{0.9}) & 91.8 (\textit{0.9})
		\tabularnewline 
		\hline 
		\centering BrainPedia & Default algorithm & 5-layers & 500 & 64 & 67.1 (\textit{1.7}) & 61.0 (\textit{1.6})
		\tabularnewline
		\centering & Pre-trained CAE & 5-layers & 200 & 64 & 73.8 (\textit{2.7}) & 70.0 (\textit{2.3})
		\end{tabular}
\end{table*}}

\subsubsection{Choice of hyperparameters for HCP dataset}
Performance of the different models trained with the different hyperparameters can be found in Supplementary Table~S1, available at \citep{supporting_data}. 
{For the default algorithm initialization, the best model had 4 layers and was trained with a batch size of 32 for 500~epochs. This model achieved an accuracy of 90.8\% on average of the 5-folds of cross-validation. For the pre-trained CAE initialization, the best model had 5 layers and was trained with a batch size of 64 for 200 epochs (average accuracy of 91.8\%). The best hyperparameters for each type of initialization (default and pre-trained) were used in all subsequent experiments.}

\subsubsection{Choice of hyperparameters for BrainPedia dataset}
Results for all sets of hyperparameters are available in Supplementary Table S2, available at \citep{supporting_data}. 
{For the default algorithm initialization, the model who achieved the best performance had 5 layers and a batch size of 64 for 500 epochs. This model classified the BrainPedia dataset with an average accuracy of 67.1\% and an average F1-score of 61\%. 
The performance of the pre-trained CAE was the best using a 5-layer architecture, a batch size of 64 and a training time of 200 epochs.}

\subsection{Benefits of self-taught learning on a homogeneous dataset}
Table \ref{tab:tab4} summarizes the results for the different classification experiments on the HCP datasets. 

{\renewcommand{\arraystretch}{2}
\begin{table*}[ht]
\centering
\let\TPToverlap=\TPTrlap
\caption{{Classification performance on HCP datasets of models initialized with default algorithm vs with the weights of the pre-trained CAE. Mean accuracies and standard errors of the means among the 5-folds of cross-validation are shown. Paired two samples t-tests are performed between the accuracies of the 5 models obtained with cross-validation for each type of initialization. DA: Default Algorithm initialization ; PT: Pre-Training initialization.}}\label{tab:tab4}
\begin{threeparttable}[ht]
        \centering
	\footnotesize
		\begin{tabular}{c|cc|cc|cc|cc}
		\hline
		\centering \textbf{Subjects} & \multicolumn{2}{c|}{\textbf{50}} & \multicolumn{2}{c|}{\textbf{100}} & \multicolumn{2}{c|}{\textbf{200}} & \multicolumn{2}{c}{\textbf{Global \hl{(393)}}}
		\tabularnewline
            \centering {\textbf{Maps}} & \multicolumn{2}{c|}{{1150}} & \multicolumn{2}{c|}{{2300}} & \multicolumn{2}{c|}{{4590}} & \multicolumn{2}{c}{{9017}}
            \tabularnewline
		\centering \textbf{Init.} & \textbf{DA} & \textbf{PT} & \textbf{DA} & \textbf{PT} & \textbf{DA} & \textbf{PT} & \textbf{DA} & \textbf{PT}
		\tabularnewline
		\hline 
		\multicolumn{9}{c}{\textbf{Contrast classification} (\textit{23 classes})}
		\tabularnewline
		\hline
		\centering Mean Acc. (\%) & 83.6 & 87.0 & 86.8 & 89.9 & 88.6 & 90.2 & 90.9 & 92.4 
		\tabularnewline 
            \centering ({std. err.}) & (0.61) & (0.51) & (0.69) & (0.34) & (0.84) & (1.46) & (0.38) & (0.44) 
            \tabularnewline
		\hline 
            \centering Paired T-test (\textit{4 dof}) & & \textbf{-11.52} & & \textbf{-4.77} & & -1.42 & & \textbf{-4.74 }
            \tabularnewline
            \centering \textit{p-value} & & \textit{\textbf{0.0003}} & & \textit{\textbf{0.009}} & & \textit{0.23} & & \textit{\textbf{0.009}} 
            \tabularnewline
            \hline
		\multicolumn{9}{c}{\textbf{Task classification} (\textit{7 classes, multiple contrasts per class})}
		\tabularnewline
		\hline
		\centering Mean Acc. (\%) & 96.6 & 97.3 & 95.4 & 98.0 & 97.9 & 98.5 & 98.4 & 99.0 
		\tabularnewline 
            \centering ({std. err.}) & {(0.47)} & {(0.43)} & {(1.49)} & {(0.25)} & {(0.44)} & {(0.16)} & {(0.17)} & {(0.13)}
            \tabularnewline
		\hline
            \centering Paired T-test (\textit{4 dof}) & & \textbf{\textit{-3.57}} & & -1.4 & & -1.5 & & \textbf{\textit{-5.65}} 
            \tabularnewline
            \centering \textit{p-value} & & \textit{\textbf{0.02}} & & \textit{0.2} & & \textit{0.2} & & \textit{\textbf{0.005}}
            \tabularnewline
            \hline
		\multicolumn{9}{c}{\textbf{One constrast task classification} (\textit{7 classes, one contrast per class})}
		\tabularnewline
		\hline
		\centering Mean Acc. (\%) & 97.9 & 99.1 & 98.9 & 99.4 & 99.3 & 99.6 & 99.4 & 99.6
		\tabularnewline 
            \centering ({std. err.}) & {(0.3)} & {(0.3)} & {(0.17)} & {(0.25)} & {(0.2)} & {(0.2)} & {(0.2)} & {(0.14)}
            \tabularnewline
		\hline
            \centering Paired T-test (\textit{4 dof}) & & \textbf{-4.17} & & \textbf{-3.32} & & -2.33 & & -2.06 
            \tabularnewline
            \centering \textit{p-value} & & \textit{\textbf{0.01}} & & \textit{\textbf{0.03}} & & \textit{0.08} & & \textit{0.1 }
            \tabularnewline
		\end{tabular}
     \end{threeparttable}
\end{table*}
}

\begin{figure}
	\centering
		\includegraphics[width=0.5\textwidth]{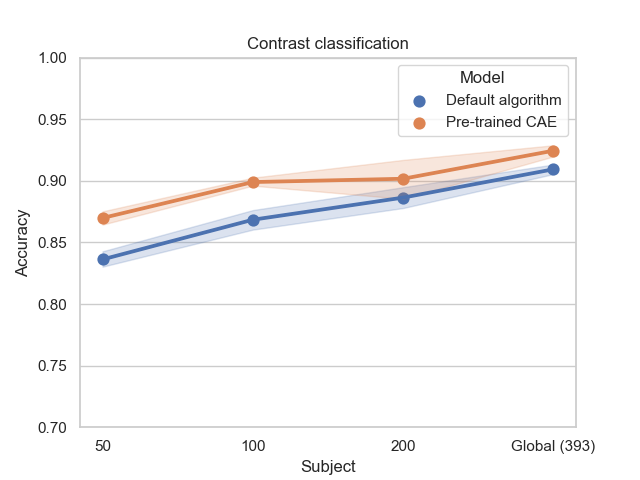}
	\caption{\hl{Mean accuracies and standard errors of the mean on} \underline{contrast classification} {with the HCP dataset for the models initialized with default algorithm (blue) and pre-trained CAE (orange).  Pre-training improves contrast classification performance for small sample sizes and at a lower level of improvement, also for large sample sizes.}}
	\label{fig:fig5}
\end{figure}

\subsubsection{Impact of the sample size}
For all classification experiments, the size of the training set (in terms of number of subjects) had a strong impact on the benefits of self-taught learning. With 50 subjects, the performance of the pre-trained CAE outperformed the performance of the classifier initialized with the default algorithm in all our experiments ({improvements of 0.7\% to 3.4\% in mean accuracies}). {These improvements were always significant (P-values < 0.05). When sample size increased, this improvement reduced and was sometimes not significant}. If we focus on contrast classification (Figure  \ref{fig:fig5}), which was the hardest classification task between the three presented here due to the higher number of classes, {the difference between the performance of the two classifiers decreased with sample size (mean accuracies of 88.6\% and 90.2\% respectively for default initialization and pre-trained model respectively for N=200 which corresponded to an improvement of 1.6\% compared to almost 3\% for N=100). For N=200, the difference of performance was not significant, probably due to the presence of an outlier value in the accuracies of the pre-trained CAE. Indeed, accuracies of the pre-trained CAE model were superior to the ones of the default model, except for the pre-trained model tested on the 3rd fold of cross-validation which was lower. This value was also significantly lower than those of models tested on other folds of cross-validation (see Supplementary Table S3, available at \citep{supporting_data}).}

\begin{figure}
	\centering
		\includegraphics[width=0.5\textwidth]{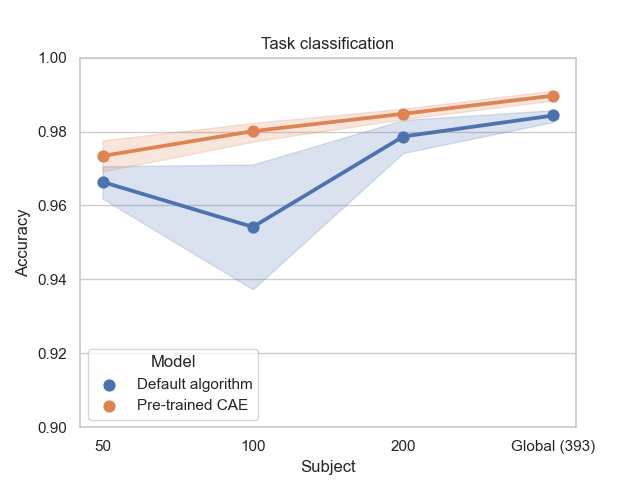}
	\caption{\hl{Mean accuracies and standard errors of the mean on} \underline{task classification} {with the HCP dataset for the models initialized with default algorithm (blue) and pre-trained CAE (orange). Pre-training improves task classification performance for all sample sizes but sample sizes did not have a huge influence on the level of improvement.}}
	\label{fig:fig6}
\end{figure}

\begin{figure}
	\centering
		\includegraphics[width=0.5\textwidth]{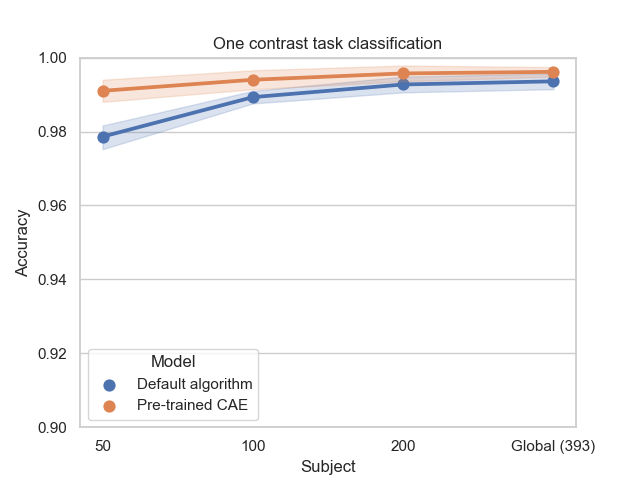}
	\caption{\hl{Mean accuracies and standard errors of the mean on} \underline{one contrast task classification} {with the HCP dataset for the models initialized with default algorithm (blue) and pre-trained CAE (orange). Pre-training does not always improve one-contrast task classification performance: for large sample sizes, pre-training and default initialization give very similar results.}}
	\label{fig:fig7}
\end{figure}

\subsubsection{Impact of the target classification task}
For simpler classification experiments (\textit{i.e.} with less classes to separate), pre-training was not always useful. In these experiments, performance was already nearly perfect (accuracies close to 1) and therefore difficult to improve. {For large sample sizes ($N > 100$), performance was close (difference between mean accuracies lower than 0.6\%) between models initialized with default algorithm and pre-trained models} (see Figures \ref{fig:fig6} and \ref{fig:fig7}). However, for smaller sample sizes (N=50), pre-training improved classification -- similarly to what had been shown for more complex tasks -- with accuracies of the pre-trained models higher than default models of 0.7\% and 1.2\% for task classification and one contrast task classification respectively. These results suggest that pre-training can be beneficial when studying difficult classification problems such as those with few training samples or complex classification tasks.

\subsection{Benefits of self-taught learning on a heterogeneous dataset}
\begin{figure}
	\centering
		\includegraphics[width=0.5\textwidth]{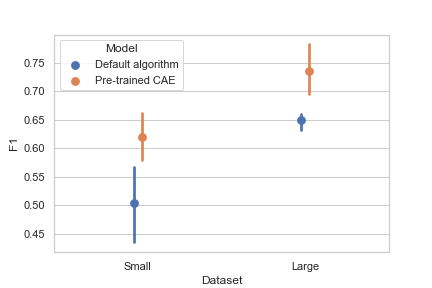}
	\caption{Mean F1-scores and standard errors of the mean of the classification of mental concepts on BrainPedia datasets (Small and Large) for the models initialized with default algorithm (blue) and pre-trained CAE (orange). Pre-training improves classification performance, in particular for the small dataset.}
	\label{fig:fig8}
\end{figure}

Table \ref{tab:tab5} summarizes the results for the classification of mental concepts on the small and the large BrainPedia datasets. These results are illustrated in Figure \ref{fig:fig8}.

{\renewcommand{\arraystretch}{2}
\begin{table*}[h]
	\centering
	\caption{{Classification performance on BrainPedia datasets of models initialized with default algorithm vs with the weights of a pre-trained CAE. DA: Default Algorithm initialization ; PT: Pre-Training initialization}} \label{tab:tab5}
 \begin{threeparttable}[ht]
	\footnotesize
		\begin{tabular}{c|cc|cc}
		\centering \textbf{Dataset} & \multicolumn{2}{c|}{\textbf{Small BrainPedia}} & \multicolumn{2}{c}{\textbf{BrainPedia}}
		\tabularnewline
		\centering \textbf{Init.} & \textbf{DA} & \textbf{PT} & \textbf{DA} & \textbf{PT}
		\tabularnewline
		\hline
		\centering \textbf{Mean acc. (\%)} & 56.8 & 64.5 & 67.1 & 74.2
		\tabularnewline
            \centering {(std. err.)} & {(1.5)} & {(2.1)} & {(0.9)} & {(2.3)}
		\tabularnewline
		\hline
            \centering Paired T-test (\textit{4 dof}) & & \textbf{-8.72} & & \textbf{-3.43}
            \tabularnewline
            \centering \textit{p-value} & & \textit{\textbf{0.001}} & & \textit{\textbf{0.02}}
            \tabularnewline
            \hline
		\centering \textbf{Mean F1-score (\%)} & 50.5 & 62.0 & 64.9 & 73.6
		\tabularnewline
            \centering {(std. err.)} & {(3.5)} & {(2.1)} & {(0.8)} & {(2.2)}
		\tabularnewline
            \hline
            \centering Paired T-test (\textit{4 dof}) & & \textbf{-4.89} & & \textbf{-2.89}
            \tabularnewline
            \centering \textit{p-value} & & \textit{\textbf{0.008}} & & \textit{\textbf{0.04}}
		\end{tabular}
  \end{threeparttable}
\end{table*}}

On a the small BrainPedia dataset, pre-training improved the performance of the classifier. {When looking at the mean accuracies, respectively 56.8\% and 64.5\% for the classifier initialized with the default algorithm and the pre-trained classifier, the difference was high (almost 8\% of improvement)}. But in this case, the F1-score was a better metric to assess the performance. Indeed, this metric focuses more on classification errors and is a better indicator of performance when classes are imbalanced, which was the case in this dataset in which some classes were more represented than others (\textit{e.g.} in the small BrainPedia training set, 205 maps corresponded to the class "visual words, language, visual" whereas only 19 are in the class "left foot, visual"). When focusing on this metric, the pre-trained classifier performance was markedly higher than the ones of the classifier with default initialization {(11.5\% of improvement in mean F1-score)}. {Performance (accuracies and F1-scores) was both significantly improved with the pre-trained model compared to the default one ($p < 0.05$).}

On the global BrainPedia dataset, performance also increased with pre-training. Mean accuracy and F1-score were higher for the the pre-trained model ({F1-score of 73.6\% against 64.9\% for the model with default initialization}) even if the sample size of the dataset was higher and more classes were represented. Indeed, the classification task was also more complex for this dataset since data were separated into 36 classes instead of 30 for Small BrainPedia due to the presence of maps from other studies in the dataset. 

\subsection{How do we explain these benefits?}

\subsubsection{Features}

\begin{figure}
	\centering
		\includegraphics[width=0.99\textwidth]{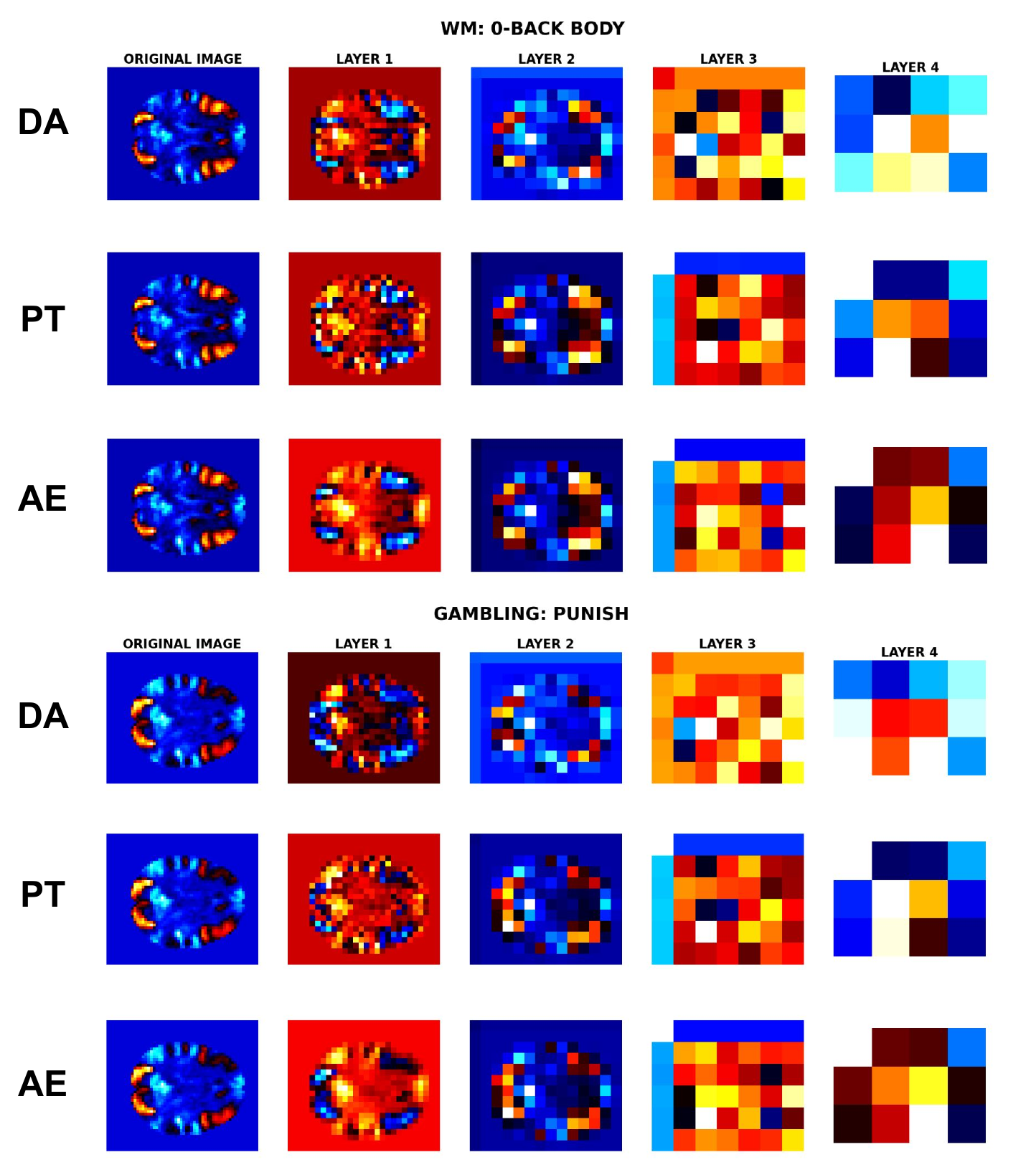}
	\caption{{Original mean statistic maps (column 1) and mean feature maps across subjects of the fold 1 of the test dataset of HCP 50 for the first four convolutional layers of each model (columns 2-5): CNN with default algorithm initialization (DA), pre-trained CNN (PT) and CAE, for two of the eight selected contrasts (WM: 0-back body and Gambling: Punish).}}
	\label{fig:fig9}
\end{figure}

{\renewcommand{\arraystretch}{2}
\begin{table}[]
    \centering
    \begin{tabular}{c|c|c}
         \centering \textbf{Contrast} & \multicolumn{2}{c}{\centering \textbf{Per-class accuracy}} 
         \tabularnewline
         \centering & \centering \textbf{DA} & \centering \textbf{PT}
         \tabularnewline
         \hline
         \centering \textbf{WM: 0BK BODY} & \centering 57.7 & \centering 60.3
         \tabularnewline
         \centering \textbf{WM: 0BK PLACE} & \centering 70.5 & \centering 79.5
         \tabularnewline
         \centering \textbf{WM: 0BK TOOL} & \centering 57.8 & \centering 66.7
         \tabularnewline
         \centering \textbf{WM: 2BK BODY} & \centering 74.3 & \centering 73.1
         \tabularnewline
         \centering \textbf{WM: 2BK TOOL} & \centering 47.4 & \centering 60.3
         \tabularnewline
         \centering \textbf{GAMBLING: PUNISH} & \centering 55.1 & \centering 67.9
         \tabularnewline
         \centering \textbf{RELATIONAL} & \centering 58.9 & \centering 75.6
         \tabularnewline
         \centering \textbf{GAMBLING: REWARD} & \centering 57.7 & \centering 66.7
    \end{tabular}
    \caption{Per-class accuracies for classification of contrasts with HCP dataset sample N=50 for DA (Default Algorithm) and PT (Pre-trained CAE). Only lowest per-class accuracy (< 80\%) are shown in the Figure. For other per-class accuracy, please refer to Supplementary Table S7, available at \citep{supporting_data}}
    \label{tab:tab6}
\end{table}}

{To better understand the behaviour of each model -- in particular on what features they based their predictions on -- we visualized the mean features across subjects of each layer of the pre-trained, default models and baseline CAE for each class label (\textit{i.e.} contrast). Specifically, we studied the mean feature maps obtained across subjects in the test set (fold 1) of the N=50 sample of the HCP dataset for different contrasts. This configuration was chosen due to the large difference between performance of default and pre-trained models on this classification task. Our main interest was to see if the model would focus on general patterns of activation or more individual features. We focused on the contrasts that led to the most difficult classification tasks (i.e had the lowest per-class accuracy (less than 80\%)). \hl{Per-class accuracy for selected contrasts are shown in Table }\ref{tab:tab6} and for all contrasts in Supplementary Table S7, available at \citep{supporting_data}. Eight contrasts were selected: `Working Memory’: `0-back body’, `0-back places’, `0-back tools’, `2-back body’, `2-back tools’ , `Gambling: punish', `Gambling: reward' and `Relational: relational' and among these 8 contrasts, 7 (all except `2-back body') had a better per-class accuracy with the pre-trained CAE,} see \ref{hcp_classif}. 

Figure \ref{fig:fig9} {shows the mean feature maps for two of the selected contrasts and for the first four convolutional layers of the models: CNN with default initialization, pre-trained CNN and CAE. The mean feature maps of all the selected contrasts and layers are displayed in Supplementary Figure S2, available at \citep{supporting_data}. The first convolutional layer features (column two of} Figure \ref{fig:fig9}) {were similar across models but different between the contrasts: see, for instance, the activation patterns of contrasts WM: `0-back body' and `Gambling: Punish', which were localised in the same areas, had different shapes. These were high level features: brain shape and main activation patterns. However, the second convolutional layer (third column) seemed to learn more important features for classification. The shape of the brain was still visible but patterns of activation were more blurry, as if they were lower resolution representations of the original statistic maps. However, features started to be different between models at this layer with some modifications of the shape of the main activation patterns between the default model (first row of each contrast) vs. the pre-trained model and the CAE (second and third lines). The same observation was made for the third convolutional layer (fourth column), which began to learn deeper representations. Due to the size of the features (6 * 7 * 6), the brain shape and activation patterns were not visible, these features were thus less interpretable and required a quantitative analysis.}

\begin{figure}[h]
	\centering
		\includegraphics[width=0.99\textwidth]{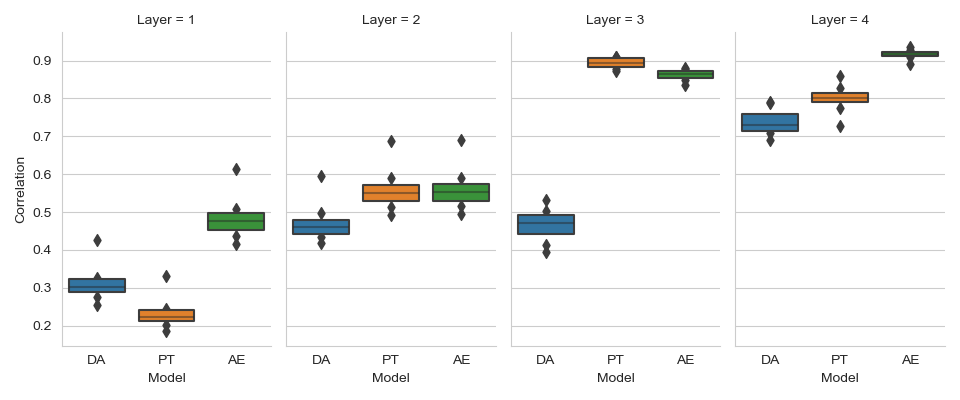}
	\caption{{Boxplots of mean correlations between the feature maps of different subjects for the eight selected contrasts (`Working Memory’: `0-back body’, `0-back places’, `0-back tools’, `2-back body’, `2-back tools’ , `Gambling: punish', `Gambling: reward' and `Relational: relational') for different models at layer 1, 2, 3 and 4. DA: Default Algorithm initialization ; PT: Pre-Training initialization ; AE: Baseline AutoEncoder. For Layers 3 and 4, pre-trained CNN and baseline CAE show larger correlation between subjects than default CNN, meaning a lower attention to individual variabilities.}}
	\label{fig:fig10}
\end{figure}

{Mean correlations between the feature maps of the same contrast were computed for each pair of subjects. A high mean correlation indicates a higher similarity between the feature maps produced in a given layer of a neural network, and thus potentially, a higher generalisation power since the feature maps are less different between subjects and thus less sensitive to individual variations.} Figure \ref{fig:fig10} {shows the mean correlations for the 8 selected contrasts and for the first four convolutional layers of the models (different values represent different contrasts). For layers 1 and 2, mean correlations were low (<60\%) and not very different between the models even if the pre-trained CNN seemed to account more about individual differences than the default model and baseline CAE. The main change was visible at layer 3 where there was an important difference (more than 30\% for every contrast) between the mean correlation between the features learned by the default CNN and the pre-trained one. The features of this layer seemed more similar between different subjects and more generalizable across subjects for the pre-trained model (mean correlations>80\% for all contrasts) than for the default model for which the mean correlations were lower than 50\% for every contrast. Correlations started to converge for the fourth layer, but were still lower for the default model.}

\subsubsection{What layers benefit the most from weight transfer from the CAE?}

{\renewcommand{\arraystretch}{2}
\begin{table}[]
    \centering
    \begin{tabular}{c|c}
         \centering \textbf{Number of transferred layers} & \centering \textbf{Mean classification accuracy (standard error) (\%)} 
         \tabularnewline
         \hline
         \centering \hl{\textbf{0 (Default initialization)}} & \hl{83.6 (0.61)}
         \tabularnewline
         \centering \textbf{1} & 82.67 (0.45) 
         \tabularnewline
         \centering \textbf{2} & 84.79 (0.52) 
         \tabularnewline
         \centering \textbf{3} & 85.51 (0.8) 
         \tabularnewline
         \centering \textbf{4} & 86.6 (0.4) 
         \tabularnewline
         \centering \textbf{Full pre-trained model} & 87.0 (0.51)
    \end{tabular}
    \caption{Classification performance (mean accuracy and standard error, in \%) of pre-trained models with different numbers of transferred layers on classification of contrasts for HCP dataset sample n=50.}
    \label{tab:tab7}
\end{table}}

{To explore the impact of each layer and the benefits of the baseline weights of the CAE, we tried several experiments with different numbers of frozen layers and several weight transfer configurations: transferring only the weights of the first convolutional layer to transferring the weights of the first four convolutional layers. Performance of the different models with different numbers of transferred layers is shown in Table} \ref{tab:tab7}{. When only the weights of the first layer were transferred, classification performance was lower than with other configurations (82.7\% of accuracy compared to more than 84\% for at least 2 transferred layers). This suggests that features learned by the CAE at this layer were less important for classification. However, when increasing the number of transferred layers, performance started to grow and became closer to the accuracy obtained when transferring all layers (87\%). This growth was quite constant and there was no large improvement of performance when transferring the weights of a layer in particular, except when moving from transferring the first layer to the first two layers. Thus, pre-training the deeper layers of the model was beneficial to improve classification performance, probably because of the ability of these layers to extract more general features, less sensitive to individual variations, as we saw above. Transferring the weights of the last convolutional layer (5th) was however not very impactful, performance of model with four transferred layers was very close to the ones of fully pre-trained model (86.6\% vs 87.0\%). We suppose that this layer was important to extract task-related features that were different from the ones learned by the CAE, explaining the limited impacts of transferring the CAE weights.}

\subsubsection{Faster fine-tuning: what happens if we freeze some layers?}

{\renewcommand{\arraystretch}{2}
\begin{table}[]
    \centering
    \begin{tabular}{c|c}
         \centering \textbf{Nb. of frozen layers} & \centering \textbf{Mean classification accuracy (standard error) (\%)} 
         \tabularnewline
         \hline
         \centering \textbf{2} & 86.7 (0.54) 
         \tabularnewline
         \centering \textbf{3} & 86.82 (0.66)
         \tabularnewline
         \centering \textbf{4} & 86.1 (0.64) 
         \tabularnewline
         \centering \textbf{5} & 80.42 (0.99)
    \end{tabular}
    \caption{Classification performance (mean accuracy and standard error, in \%) of pre-trained models with different numbers of frozen layers on classification of contrasts for HCP dataset sample n=50.}
    \label{tab:tab8}
\end{table}}

Table \ref{tab:tab8} {shows the results of the different experiments with different numbers of frozen layers. When we froze the first convolutional layers (from 2 to 4 frozen layers) on the pre-trained model, the performance did not decrease. This suggests that the features extracted by the baseline autoencoder for these layers were general enough to perform a classification task with only one fine-tuned convolutional layer in addition to the dense layer. However, when freezing all convolutional layers of the model (5 layers), there was a large drop in terms of performance (86 to 80\% of accuracy between freezing 2-4 layers vs 5 layers), this confirmed the observation made before on the difference between the features extracted by the fifth layer for reconstruction (CAE) and for classification (CNN). In conclusion, the first four convolutional layers of our model extracted more general features whereas the last one extracted deeper and more specific features for classification.}

\section{Discussion}
\subsection{Summary of the results}
In this work, we showed the benefits of self-taught learning with a large and variable database on the classification of two large public datasets with different sample sizes and classification tasks. In all cases, pre-training a classifier with an unsupervised task (in our case: reconstruction) was beneficial but the level of improvement varied depending on the classification task and the size of the training dataset.

When sample sizes were small, pre-training always improved the classification performance, regardless of whether the dataset was homogeneous or heterogeneous and of the complexity of the classification task. In medical imaging, where the dimensions of the data are often very large and few samples are typically available due to high financial and human costs, learning a good representation of the data can be very difficult \citep{thomas_challenges_2021, Sun_2017_ICCV}. Unsupervised pre-training can thus be helpful by initializing the weights of the CNN to preserve the (brain) structure learned by the autoencoder,
and facilitate the learning process. However, when the sample size increases, benefits are less remarkable since the amount of available training data is probably sufficient to learn a good representation. 

This observation can also be made for classification tasks. When trying to classify the data in a small number of classes, performance of the pre-trained classifier was better but not with a high improvement of performance, even for small sample sizes (\textit{e.g.} 100 subjects for task classification). But when trying to separate data into more classes, for a more fine-grained classification, the representation learned during the pre-training was beneficial. 

Another benefit of self-taught learning we found was the reduction of the training time. Performance of the pre-trained classifier was better even with less training epochs. This was the case for both datasets results which were computed for 500 epochs for the default algorithm and 200 epochs for the pre-trained model. This is in line with~\cite{neyshabur2020being} in which researchers showed that the pre-trained models remain in the same basin of the loss function when trained on new data and since the weights are already initialized close to a good representation of data, less epochs are necessary to adapt this representation for classification. 

Architectures of the models also had an impact on the benefits of self-taught learning. With both datasets, pre-trained models performed better using the 5-layers architecture. This effect was studied by \cite{erhan_why_nodate} who showed that, while
unsupervised pre-training helps for deep networks with more layers, it appears to hurt for too small networks. The size of the latent space of the CAE with 5-layers being almost 5 times smaller that the 4-layers one, it suggests that only a small subset of features of the input are relevant for predicting the class label. 

However, the classification accuracies of the pre-trained models were not related to the reconstruction performance of the CAE since the 4-layers CAE reconstructs maps with better precision than the 5-layers CAE. This confirms that the features learned by the 4-layers CAE for reconstruction were not all useful for classification and focusing on a smaller number of features (with 5-layers) facilitates the learning process. 

{This observation was confirmed by the large drop in performance when freezing the first fifth convolutional layers of the pre-trained model and when transferring only part of the layers. Deeper pre-trained layers had more impact on classification performance, meaning that the features extracted by these layers were different from those learned by layers initialized with the default algorithm. In particular, the third and fourth convolutional layers showed the best benefits when being transferred, due to the generalizability of the extracted features. This was not the case for the fifth layer, for which features need to be specific to the classification task. }

{The pre-trained model improved the performance in terms of classification due its ability to focus on more generalizable features. By pre-training a model on a large variable dataset such as NeuroVault, we built a model that is less sensitive to the training data and less sensitive to individual differences, thus more generalizable and applicable to new subjects.}

\subsection{Limitations and perpectives}
Due to the high computational time required to train a model, we only compared two model architectures (4 and 5-layers). {Indeed, training a CAE model can be very time consuming, particularly in our case since we use a large training dataset (N=22,772) and high dimensional data (k=48 * 56 * 48). With the 4-layers model, for 200 epochs it took approximately 48h to train on 1 GPU. With parallel computing (use of 2 GPUs in parallel), we could hope to shorten this time to 24h, with the cost of using more computing resources.} Other types of architectures with different number of fully-connected or convolutional layers could have been tested to see the effect of other latent space sizes as it was done in \cite{erhan_why_nodate}.

The main limitation of our work is the classification experiments and datasets we chose. In fMRI, the number of possible labels and thus, classification tasks is very high due to a lack of consensus in the field with respect to standardizing tasks, contrasts and mental concepts~\citep{poldrack_cognitive_2011}. In our experiments, we used the labels provided by NeuroVault as specified in the original studies \citep{van_essen_wu-minn_2013, varoquaux_atlases_2018}. We chose to compare multiple types of classification on the HCP dataset to illustrate different approaches are in use in the field or that were used by other studies \citep{y_gao_decoding_2019, thomas_evaluating_2021}. For BrainPedia, a multi-label decoding was performed in the original study since multiple concepts are associated with most maps. Labels we had access to were then the list of labels associated with each map. To be able to compare our results with those of the homogeneous dataset (HCP), we chose to classify these as unique labels, which was less complex and less precise in practice. This type of issue is due to the lack of harmonization in the way tasks and cognitive processes are defined. Using ontologies such as Cognitive Atlas \cite{poldrack_cognitive_2011}, NeuroVault annotations could be harmonized and enriched, as it was done by \cite{menuet_comprehensive_2022} by mapping the original labels to target ones from Cognitive Atlas or \cite{WALTERS2022119610} in which cognitive conditions were annotated by a group of expert using the same atlas. 

In neuroimaging, many sources of variability can impact the results of an experiment and the generalizability of the results. Here, we investigated the generalizability of our model by assessing the benefits of pre-training on a heterogeneous dataset (BrainPedia). While this dataset was heterogeneous in terms of the studies that were included,  all maps were obtained using the same processing pipeline. Multiple studies have shown that the exact pipeline used to obtain an fMRI result can have a non-negligeable impact on fMRI statistic maps~\citep{carp_plurality_2012, botvinik2020variability}. In the future, investigating performance of classification on a more variable target dataset with statistic maps from different studies but also processed using different pipelines would be of great interest. In a recent study~\cite{vu_fmri_2020}, the authors tried to compare the performance of different classifiers trained on fMRI 3D volumes series obtained with various scenarios of minimal preprocessing pipelines. {A similar experiment was recently made by} \cite{pipeline_invariant} {who found that preprocessing pipeline selection can impact the performance of a supervised classifier.} Comparing the adaptation capacities of models on volumes preprocessed with different pipelines could be also interesting to evaluate the impact of analytical variability on deep learning with fMRI {and to see if the generalizability of our pre-trained models also works for inter-pipeline differences}.

{Note that self-supervised (instead of self-taught) learning could have also been used to pre-train our model, as it was done by}\cite{self_sup}{ who designed self-supervised learning frameworks, inspired by the field of natural language processing, to pre-train mental state decoding models. Self-supervised learning is a supervised machine learning setting where the supervision is generated directly from the data and the model is pre-trained using a supervised surrogate task. Self-supervised is particularly relevant if the surrogate task is close to the final one targeted by the user, e.g. if they can share the same feature representation. It is possible that, by designing a relevant supervised surrogate task that could be relevant for all very diverse usage of our model, the pre-trained model would have performed better than the one presented in this article. Designing and experimenting with such a surrogate supervised task could be interesting for future work}.

In our self-taught context, using unsupervised models could allow us to build a space capturing the similarities and differences of statistic maps, \textit{i.e.} to learn a robust latent representation of the important features of statistic maps in a specific context. By adding other constraints to this latent space and/or choosing an adapted pre-training dataset, we could use this for other purposes than brain decoding. For instance, building a space that captures the analytical variability in statistic maps could help us understand the difference between the pipelines but also identify the more robust pipelines. Future works will focus on building such a space with specific constraints to evaluate distance between different pipelines. 

\section{Conclusion}
In this study, we compared the benefits of a self-taught learning framework in the task of classifying 3D fMRI statistic maps. We showed that unsupervised pre-training improves the performance of classification in experiments with few training samples and complex classification tasks, which is a very common setup in fMRI studies. 

\section{Acknowledgements}
This work was partially funded by Region Bretagne (ARED MAPIS) and Agence Nationale pour la Recherche for the programm of doctoral contracts in artificial intelligence (project ANR-20-THIA-0018).
We thank Gregory Kiar who worked on a preliminary version of the autoencoder based on NeuroVault data.
Data used in the HCP dataset were provided [in part] by the Human Connectome Project, WU-Minn Consortium (Principal Investigators: David Van Essen and Kamil Ugurbil; 1U54MH091657) funded by the 16 NIH Institutes and Centers that support the NIH Blueprint for Neuroscience Research; and by the McDonnell Center for Systems Neuroscience at Washington University.

\section{Availability of data and materials}
Supporting data, including supplementary materials (figures and tables) and additional links (code and derived data) is also available via the GigaScience repository, GigaDB \cite{supporting_data}.

\subsection{Data availability}
The data used in this study are openly available on NeuroVault \cite{gorgolewski_neurovaultorg_2015}. No experimental activity involving the human participants was made by the authors. Only publicly released data were used.

NeuroVault IDs of statistic maps included in each dataset (NeuroVault, HCP and BrainPedia) are available in Derived data, see \ref{derived}. 

\subsection{Code}
The code produced to run the experiments and to create the figures and tables of this paper is available in the Software Heritage public archive at \cite{code_SH}. 
\begin{itemize}
    \item Programming language: Python3.9
    \item Licence: MIT 
    \item Requirements: numerous Python libraries described in the GitLab repository.
\end{itemize}

\subsection{Derived data} \label{derived}
Derived data such as NeuroVault IDs of statistic maps used in the different datasets and the parameters of the trained models (for validation of hyperparameter and test of performance) are available in Zenodo: \cite{derived_data}.

\subsection{Ethics}

The data used in this study are openly available on NeuroVault \cite{gorgolewski_neurovaultorg_2015}. No experimental activity involving the human participants was made by the authors. Only publicly released data were used.

\textbf{HCP}: Written informed consent was obtained from all participants of HCP and the original study was approved by the Washington University Institutional Review Board. 

\textbf{BrainPedia}: BrainPedia database is comprised of data from 29 studies, assembled from various sources (OpenNeuro, NeuroSpin research center, etc.). Subject-level maps resulting from first-level statistical analysis were upladed on NeuroVault by \cite{varoquaux_atlases_2018}. The list of the original dataset used is available on \cite{varoquaux_atlases_2018}, S1 Text \cite{supp_mat_brainpedia}.

\bibliography{references}  

\end{document}